\documentclass[a4paper,fleqn]{cas-dc}

\usepackage{algorithm2e}
\usepackage{adjustbox}
\usepackage{natbib}
\usepackage{xcolor}

\usepackage{}

\def\tsc#1{\csdef{#1}{\textsc{\lowercase{#1}}\xspace}}
\tsc{WGM}
\tsc{QE}
\tsc{EP}
\tsc{PMS}
\tsc{BEC}
\tsc{DE}

\begin{document}
\let\WriteBookmarks\relax
\def\floatpagepagefraction{1}
\def\textpagefraction{.001}
\shorttitle{Exponential Moving Average Guided Knowledge Distillation for Robust Federated Learning}
\shortauthors{H. Reguieg, M. El Kamili, E. Sabir}

\title [mode = title]{FedEMA-Distill: Exponential Moving Average Guided Knowledge Distillation for Robust Federated Learning}\tnotemark[1]

\tnotetext[1]{This work was supported by TELUQ university under research grant No. 711823.}
   



\author[1,2]{Hamza Reguieg}[orcid=0009-0004-3186-3527]
\cormark[1]
\ead{hamza.reguieg@teluq.ca; hamza.reguieg-etu@etu.univh2c.ma}
\credit{Conceptualization, Methodology, Formal analysis, Software, Writing -- Original Draft}
\cortext[1]{Corresponding Authors: hamza.reguieg@teluq.ca, hamza.reguieg-etu@etu.univh2c.ma (H. Reguieg); essaid.sabir@teluq.ca (E. Sabir)}

\author[2]{Mohamed {El Kamili}}[orcid=0000-0001-8039-684X]
\ead{mohamed.elkamili@etu.univh2c.ma}
\credit{Supervision, Validation, Writing, Review \& Editing}

\affiliation[1]{organization={Department of Science and Technology, TÉLUQ, University of Quebec, H2S 3L4},
                city={Montreal},
                country={Canada}}
\affiliation[2]{organization={Higher School of Technology, Hassan II University},
  city={Casablanca}, country={Morocco}}

\author[1]{Essaid Sabir}[orcid=0000-0001-9946-5761]
\cormark[1]
\ead{essaid.sabir@teluq.ca}
\credit{Validation, Writing, Review \& Editing, Funding acquisition}



\begin{abstract}
Federated learning (FL) often degrades when clients hold heterogeneous non-Independent and Identically Distributed (non-IID) data and when some clients behave adversarially, leading to client drift, slow convergence, and high communication overhead. This paper proposes FedEMA–Distill, a server-side procedure that combines an exponential moving average (EMA) of the global model with ensemble knowledge distillation from client-uploaded prediction logits evaluated on a small public proxy dataset. Clients run standard local training, upload only compressed logits, and may use different model architectures, so no changes are required to client-side software while still supporting model heterogeneity across devices. Experiments on CIFAR-10, CIFAR-100, FEMNIST, and AG News under Dirichlet-0.1 label skew show that FedEMA–Distill improves top-1 accuracy by several percentage points (up to +5\% on CIFAR-10 and +6\% on CIFAR-100) over representative baselines, reaches a given target accuracy in 30–35\% fewer communication rounds, and reduces per-round client uplink payloads to 0.09–0.46 MB, i.e., roughly an order of magnitude less than transmitting full model weights. Using coordinate-wise median or trimmed-mean aggregation of logits at the server further stabilizes training in the presence of up to 10–20\% Byzantine clients and yields well-calibrated predictions under attack. These results indicate that coupling temporal smoothing with logits-only aggregation provides a communication-efficient and attack-resilient FL pipeline that is deployment-friendly and compatible with secure aggregation and differential privacy, since only aggregated or obfuscated model outputs are exchanged.


\end{abstract}


\maketitle

\section{Introduction}
Federated learning (FL) enables multiple data owners to collaboratively train a shared model without exchanging raw data \citep{mcmahan17}. However, two key challenges hinder practical FL deployments: (i) data heterogeneity clients often have \emph{Independent and Identically Distributed} (non-IID) datasets (e.g., users have personal or biased datasets), which induces biased local model updates (a phenomenon known as \emph{client drift}) and can significantly degrade the global model’s accuracy \citep{hsu19dirichlet} and (ii) communication constraints frequent transmission of high-dimensional model updates (often tens of megabytes) can be infeasible for mobile/edge devices with limited bandwidth or power. Addressing both stability under non-IID conditions and communication efficiency remains a central challenge for FL.

\noindent In real-world deployments (e.g., mobile phones, healthcare networks, finance), these issues appear together: clients have biased data (causing drift and slow or unstable convergence of standard FedAvg \citep{mcmahan17} and tight uplink budgets (making it costly to send full model weights each round). Biased local updates can cause the global model to oscillate or even diverge if not properly controlled, while large model updates strain network resources. Two complementary lines of work have emerged to tackle these problems:

\noindent \textbf{(i) Reducing drift and variance:} Methods modify local or server optimization to stabilize training on non-IID data. For example, \emph{FedProx} adds a proximal term to keep local updates closer to the global model \citep{li20fedprox}; \emph{SCAFFOLD} uses control variates to correct client drift \citep{karimireddy20scaffold}; the \emph{FedOpt} family (e.g., FedAvgM, FedAdam) applies momentum or adaptive learning rates on the server to smooth oscillations \citep{reddi21fedopt}; and \emph{FedDyn} introduces a dynamic regularizer per client to account for heterogeneity. These approaches improve convergence on non-IID data, but they still rely on exchanging full model weights or gradients — thus not directly addressing communication cost or client architecture differences.Moreover, improving stability typically reduces the number of communication rounds needed to reach a target accuracy; thus, even when per-round payloads remain weight-level, drift reduction can still lower total end-to-end communication and device energy by converging in fewer rounds.

\noindent \textbf{(ii) Output-level aggregation and knowledge distillation:} A separate thread focuses on communication efficiency (and allowing heterogeneous models) by sharing model outputs (predictions) instead of weights. Clients compute logits or soft labels on a public dataset and send those to the server. The server then aggregates these outputs and uses them to update the global model via knowledge distillation. Representative methods include \emph{FedMD} \citep{fedmd} and \emph{FedDistill}, which perform client-to-server knowledge transfer using a public dataset; \emph{FedDF} \citep{lin20feddf}, where the server ensembles client models’ logits on unlabeled data to guide the global model and \emph{FedBE} \citep{chen20fedbe}, which improves FedDF by Bayesian ensembling to produce better calibrated “teacher” logits. These distillation-based pipelines drastically reduce uplink communication and naturally support model heterogeneity (clients and server can have different architectures, as long as they share a common output label space). When distillation-only training exhibits higher round-to-round variance under strong label skew, it may require additional rounds to stabilize, which can reduce the practical advantage in total communication despite low uplink per round. However, prior work has noted that these methods can suffer from temporal inconsistency or higher round-to-round variance, especially under strong data skew \citep{lin20feddf,chen20fedbe}. Without a mechanism to carry over “momentum” from round to round, the global model in distillation approaches might oscillate as it is re-trained from scratch each round.

\noindent Complementary empirical studies, including our comparative evaluation of FedAvg and Per-FedAvg under Dirichlet-controlled heterogeneity and a coalition-based FL scheme for IoT devices, further confirm that strong label skew and constrained uplinks jointly exacerbate client drift and communication costs for standard aggregation rules \citep{reguieg2023comparative,elhanjriCoalitionFL}. In summary, drift-reduction methods provide temporal stability but don’t reduce communication, while output-distillation methods save communication but can introduce instability in non-IID settings. There is a gap in jointly addressing both facets.\\

\noindent In this work, we present \textsc{FedEMA-Distill}, a server-centric FL algorithm that couples temporal smoothing with logit-based aggregation. In each round, clients train locally and instead of sending model weights, they send only their model’s output probabilities (logits) on a small public proxy dataset. The server aggregates these logits (optionally using robust statistics), performs a brief knowledge distillation (KD) step to update the global model, and then applies an Exponential Moving Average (EMA) to the global model weights before the next round. Table~\ref{tab:comparison} contrasts our approach with standard FedAvg and pure distillation.

\begin{table}[t]
\centering
\caption{Comparison of FL aggregation methods. FedEMA-Distill combines the strengths of weight-averaging and logits distillation: small client uploads and heterogeneity support (like FedDF), plus improved stability and robustness (via EMA and robust aggregation).}
\label{tab:comparison}
\begin{tabular}{@{}p{2.2cm}p{2.6cm}p{2.6cm}@{}}
\toprule
\textbf{Aspect} & \textbf{FedAvg/FedProx (weights)} & \textbf{FedDF / FedBE (logits)} \\
\midrule
Client upload/round & Full model ($\sim$3.8 MB) & Logits (0.1--0.2 MB) \\
Server broadcast/round & Full model & Full model \\
Supports model heterogeneity & No & Yes \\
Robustness to Byzantine clients & Limited (robust weight avg) & Limited (no built-in robust) \\
Stability on non-IID dataset & Moderate (improved by momentum) & Low (per-round re-training variance) \\
Server compute/round & Low (averaging) & Medium (distillation) \\
\midrule
FedEMA--Distill & \multicolumn{2}{@{}p{5.2cm}@{}}{Logits + EMA; low uploads; heterogeneity; robust logit aggregation; high stability.} \\
\bottomrule
\end{tabular}
\end{table}

Contributions: (1) \emph{EMA-guided server-side distillation (zero client-side changes):} a new FL algorithm where the server maintains an EMA of the global model while aggregating only client logits\\ (2) \emph{Improved efficiency and stability under heterogeneity:} higher accuracy and faster convergence (fewer rounds) with orders-of-magnitude lower uplink\\ (3) \emph{Robustness to adversarial clients at the logit level:} coordinate-wise median/trimmed-mean secure training with up to $\sim$20\% Byzantine clients\\ (4) \emph{System perspective:} compatibility with secure aggregation \citep{bonawitz17secureagg} and differential privacy, alongside an energy analysis for edge devices.\\

\noindent\textbf{Experimental highlights.}
 Across four benchmarks (CIFAR-10/100, FEMNIST, and AG News) under Dirichlet-0.1 label skew, FedEMA--Distill achieves the best final accuracy among the considered baselines (Table~\ref{tab:finalacc}). On CIFAR-10, it reaches 70\% accuracy in about 40 rounds (vs. 60 rounds for FedAvg) while maintaining an uplink of only 0.09MB per client per round (Table~\ref{tab:c10}). As a result, the total client upload to reach 70\% drops to about 3.6MB per client compared to about 228MB for FedAvg, corresponding to a 63$\times$ reduction (Table~\ref{tab:comm70}). These results support the core claim that coupling temporal smoothing (EMA) with logits-only aggregation improves stability under non-IID drift while simultaneously improving practical communication efficiency via faster convergence.

\section{Related Work}
Federated learning trains a global model across decentralized data while keeping raw data local\citep{mcmahan17}. Two practical challenges dominate real deployments: (i) \emph{statistical heterogeneity} (non-IID data) that induces client drift and destabilizes optimization, and (ii) \emph{communication constraints}, especially on the uplink. The first line of work reduces drift or variance by modifying client or server optimization. \emph{FedProx} constrains local objectives with a proximal term\citep{li20fedprox}; \emph{SCAFFOLD} uses control variates to correct client–server mismatch\citep{karimireddy20scaffold}; the \emph{FedOpt} family interprets aggregation as a server optimizer (momentum and adaptive variants)\citep{reddi21fedopt}; \emph{FedDyn} recenters local objectives via dynamic regularization\citep{acaretal21feddyn}. Complementing these methods, our comparative evaluation of FedAvg and Per-FedAvg under Dirichlet-distributed client data quantifies how the Dirichlet concentration parameter and the number of local steps jointly affect accuracy and stability in non-IID regimes \citep{reguieg2023comparative}.These methods, however, retain weight-level communication.

A complementary thread performs \emph{output-level} aggregation via knowledge distillation (KD), which can cut uplink and naturally support heterogeneous client models. Early systems such as \emph{FedMD}/\emph{FedDistill} exchange logits on a small public proxy and ensemble them server-side\citep{jeong18fedmd}. \emph{FedDF} distills an ensemble of client models on server-side proxy data\citep{lin20feddf}, and \emph{FedBE} uses Bayesian ensembling to improve calibration before distillation\citep{chen20fedbe}. Communication can be reduced further by quantizing or sparsifying logits\citep{wu22fedkd}. Recent surveys systematize this design space and its choices (proxy selection, temperature, aggregation, personalization)\citep{wu2023kdSurvey,li2024fedsurvey,mora2024practicalKD}.

Within KD-based FL, there is growing interest in \emph{temporal consistency} and \emph{teacher design}. \emph{FedGKD} aggregates predictions from a history of global models to guide local updates, explicitly injecting memory at the output level\citep{yao2024fedgkd}. Alternative forms of knowledge beyond logits include\\
feature/prototype-level distillation (e.g., FedGPD) which improves heterogeneous FL by distilling global class\\prototypes\citep{wu2024fedgpd}. Black-box and adversarially assisted distillation (e.g., FedAL) address data heterogeneity without sharing model parameters\citep{han2023fedal}.
Orthogonal ideas that stabilize the global trajectory also appear in server-side or trajectory-aware methods that incorporate exponential moving averages (EMA) or related smoothing of the global iterate\citep{kou2025fedema,li2025fedgmt}. Our work combines the strengths of these directions by pairing \emph{logits-only aggregation} with \emph{explicit EMA smoothing} of the global weights.

Beyond generic aggregation rules, federated learning has also been embedded into system-level controllers for cyber–physical infrastructures. For instance, coalition-driven FL for IoT devices exploits weight similarity to form \\communication-efficient coalitions \citep{elhanjriCoalitionFL}; cooperative water-resource management between dams combines FedAS with cooperative game theory to coordinate water transfers under physical and safety constraints \citep{reguiegDamsFedAS}; and predictive QoS–aware power-control for URLLC models power allocation as a satisfactory game and uses a robust Banach–Picard learning scheme built on deep and federated learning concepts to meet reliability and latency targets at minimum energy cost \citep{abouzahirURLLC}. Survey work on FL for 6G further frames such designs within an AI-native resource-control stack, emphasizing stringent QoS and energy constraints as key drivers for future FL protocols \citep{benDriss2023FL6G}.

Robustness and privacy remain central. For robustness, classical weight-level defenses include coordinate-wise median, trimmed-mean, and selection rules like Krum\citep{yin18byzantine,blanchard17krum}. In the KD setting, recent analyses and methods study \emph{Byzantine resilience of distillation}, proposing attacks and defenses formulated directly at the logit level\citep{roux2025byzantinefd} and exploring relational/structural knowledge to mitigate forgetting\citep{sturluson21fedrad}. On privacy, secure aggregation prevents the server from seeing individual client messages (weights, gradients, or logits) and can be composed with distillation \citep{bonawitz17secureagg,mansouri2023soksecureagg}. Differential privacy can also be layered on logits or model updates to provide formal guarantees\citep{abadi16dp,geyer17dpfl}.Recent system-oriented defenses further investigate robustness and privacy in cloud FL deployments by combining trusted execution / confidential computing with fusion-based detection mechanisms; for example, ConfShield leverages attested confidential virtual machines and a dual-stage fusion strategy to enhance model confidentiality and strengthen resilience to poisoning in cloud environments\citep{cao2026confshield}. In parallel, application-driven FL systems in cyber-physical settings increasingly integrate privacy mechanisms (e.g., differential privacy and homomorphic encryption) with multi-source fusion and edge-side optimization, as illustrated by privacy-preserving federated multi-sensor fusion for real-time traffic management in connected vehicle networks\citep{rahmati2025ppmsf}.

\emph{Positioning.} Distillation-only pipelines achieve strong communication efficiency and model-architecture heterogeneity but may exhibit higher round-to-round variance under strong label skew. Momentum EMA-only server updates stabilize optimization but still require weight uploads. FedEMA-Distill explicitly couples \emph{logits-only aggregation} with \emph{EMA smoothing} of the global weights to deliver temporal continuity, reduced uplink, and robustness using simple robust statistics on logits, without any client-side code changes.

\begin{figure}
\centering
\includegraphics[width=\linewidth]{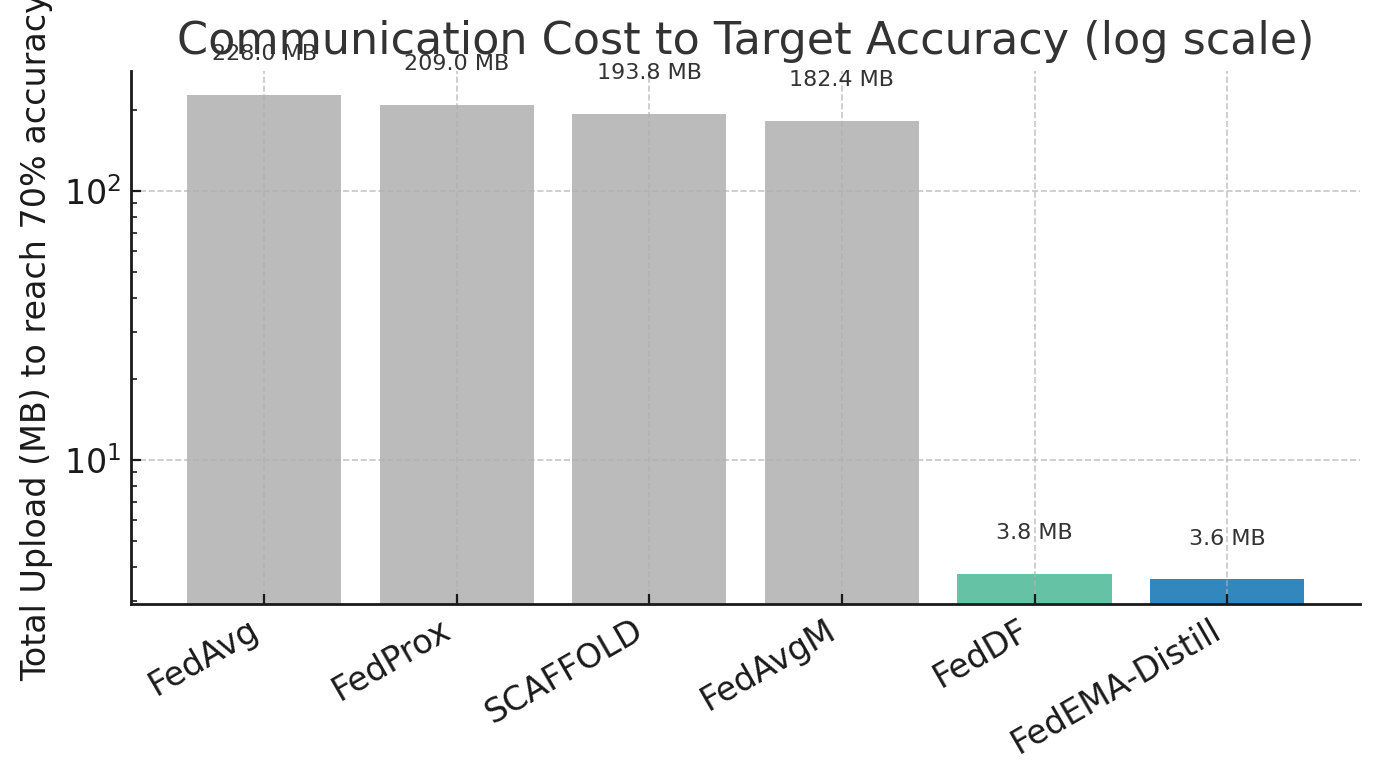}
\caption{Total upload to 70\% (log scale). Weight-based methods require $\sim$200\,MB vs. $\sim$3.6\,MB for logits-based training.}
\label{fig:comm}
\end{figure}

    \paragraph{Positioning and quantitative comparison.}
Prior FL methods addressing non-IID drift primarily operate at the \emph{weight level}, e.g., FedProx adds a proximal regularizer to stabilize local training under heterogeneity, SCAFFOLD uses control variates to correct client drift, and FedAvgM/FedOpt introduce server-side momentum or adaptive optimization to reduce oscillations. In contrast, KD-based aggregation methods such as FedDF/FedBE shift aggregation to the \emph{output level} by distilling an ensemble on a proxy set, which naturally reduces uplink and supports heterogeneous client models.
FedEMA--Distill bridges these two lines: it retains the \emph{logits-only uplink} and heterogeneity support of distillation-based FL, while injecting \emph{temporal smoothing} through server-side EMA and improving robustness via robust logit aggregation. Quantitatively, under the same non-IID protocol and identical data partitions used throughout our experiments, FedEMA--Distill consistently improves over both families. For instance, on CIFAR-10 (Dir-0.1), it outperforms drift-mitigation baselines (FedAvg/FedProx/\\SCAFFOLD/FedAvgM) in final accuracy and reaches the target accuracy in fewer rounds (Table~\ref{tab:c10}). Compared to the closest distillation baseline FedDF, which already reduces uplink, FedEMA--Distill achieves higher final accuracy and faster convergence with the \emph{same} logits uplink budget (Table~\ref{tab:c10}). Across datasets, it attains the best final accuracy (Table~\ref{tab:finalacc}). Importantly, the communication gain is substantial: to reach 70\% on CIFAR-10, weight-based baselines require on the order of hundreds of MB/client, while logits-based training requires only a few MB/client (Table~\ref{tab:comm70}, Fig.~\ref{fig:comm}). Beyond accuracy/communication, our results also indicate improved reliability under heterogeneity, including better calibration and fairer per-client performance dispersion (Sec.~\ref{sec:exp:calib-fair}).

\section{Methodology}

\subsection*{Overview}
FedEMA–Distill is a server-driven FL protocol that aggregates client model outputs (logits) on a small public proxy dataset, performs a brief server-side knowledge distillation (KD) step, and then applies an Exponential Moving Average (EMA) to the global model before broadcasting to clients. Importantly, this approach requires no changes to client-side software; clients simply compute and upload prediction logits, enabling model-architecture heterogeneity across participants. In the following, we detail the algorithm’s design, notation, and practical considerations.

\subsection*{Setting and notation.}
We consider $K$ clients with private local datasets $\{D_k\}_{k=1}^K$ (of sizes $n_k$, total $N=\sum_k n_k$). The server maintains a global model $f(\cdot; w_t)$ with weights $w_t$ at round $t$. The overall objective is to minimize the aggregate loss $F(w)$, where
$$
F(w) \;=\; \sum_{k=1}^K \frac{n_k}{N}\,F_k(w), \qquad \text{with}\;\; F_k(w)=\frac{1}{n_k}\sum_{\xi \in D_k}\ell(w; \xi)
$$
as in (1). Each client $k$ may have a different local model architecture $f_k(\cdot; \theta_{k,t})$, provided that all models share the same $C$-dimensional output space (for $C$ classes). We denote by $\mathcal{U}$ a small public proxy dataset available to the server (with $|\mathcal{U}|$ samples covering all $C$ classes). In each round $t$, the server will use a subset $\mathcal{U}_t \subseteq \mathcal{U}$ for knowledge distillation. We assume $\mathcal{U}$ is non-private (e.g. public or synthetically generated) and covers all classes. A coverage ratio $\rho \in (0,1]$ determines the fraction of proxy data used per round; over the course of training, each proxy sample will be used in roughly $1/\rho$ rounds. This helps reduce server computation while maintaining class coverage over time. The server also maintains an EMA-smoothed weight vector $\bar{w}_t$ (initialized as $\bar{w}_0 = w_0$) that accumulates the historical global model state.

\subsection*{One round of FedEMA–Distill.}
Each communication round consists of standard local training at the clients, followed by logits-only aggregation and an EMA update at the server. The steps for round $t$ are:
\begin{enumerate}
    \item \textbf{Downlink:} The server samples a set $S_t$ of clients (participation fraction $C_{\mathrm{part}}$ of $K$), and selects a proxy subset $\mathcal{U}_t \subseteq \mathcal{U}$ (of size $\rho|\mathcal{U}|$) which is partitioned evenly into $|S_t|$ shards ${\mathcal{U}_{t,k}}_{k \in S_t}$. The server broadcasts the current global model weights $w_t$ (clients with incompatible model architecture can ignore these weights and train from their own initialization) along with the proxy shard assignment to each chosen client $k\in S_t$.
    \item \textbf{Client side:} Each selected client $k$ initializes $\theta_{k,t}^{(0)} \leftarrow w_t$ if model architecture permits. The client then performs $E$ epochs of local training on its private data $D_k$ (e.g. using SGD) to obtain updated weights $\theta_{k,t}^{(E)}$. Next, for each proxy sample $x \in \mathcal{U}_{t,k}$, the client computes the logits $z_{k,t}(x) = f_k(x; \theta_{k,t}^{(E)}) \in \mathbb{R}^C$ and converts these to a probability vector (soft label) $p_{k,t}(x) = \sigma\big(z_{k,t}(x)/T\big)$, where $\sigma(\cdot)$ denotes Softmax and $T$ is a distillation temperature. The client then uploads only these FP16-quantized probability vectors $\{p_{k,t}(x): x \in \mathcal{U}_{t,k}\}$ to the server (no model weights or gradients are transmitted). This logit payload is typically very small (on the order of $C \times |\mathcal{U}_{t,k}|$ values, often a few hundred kilobytes).
    \item \textbf{Server aggregation on logits:} The server collects the logits/soft labels from all participating clients. For each proxy sample $x \in \mathcal{U}_t$, let $\mathcal{K}_t(x) = \{k \in S_t : x \in \mathcal{U}_{t,k}\}$ be the set of clients that were assigned $x$. The server computes an ensemble teacher prediction by aggregating the client probabilities for $x$:
    $$
    p_t^{\mathrm{teach}}(x) \;=\; \mathsf{Agg}\big(\{\,p_{k,t}(x) : k \in \mathcal{K}_t(x)\}\big),
    $$
    where $\mathsf{Agg}(\cdot)$ is an aggregation function applied coordinate-wise. In the simplest case, this is the average (the soft labels are averaged across clients, as in FedDF). For improved robustness, the server can use a coordinate-wise median or trimmed mean instead, which down-weights outlier values (defending against any corrupted or anomalous client logits). After aggregation, if a robust aggregator (median/trimmed-mean) was used, the resulting $p_t^{\mathrm{teach}}(x)$ is renormalized to ensure it is a valid probability distribution over classes. These robust aggregation rules are commonly used to mitigate Byzantine clients in FL \citep{yin18byzantine} and can be applied at the logit level as well \citep{roux2025byzantinefd}.
    \item \textbf{Server-side KD with anchoring:} The server now updates the global model using the aggregated “teacher” logits. Starting from the previous model weights ($u_{t+1} \leftarrow w_t$ as an initialization), the server performs a few steps of knowledge distillation on the proxy data $\mathcal{U}_t$. Specifically, it minimizes the loss
    \begin{equation}
\begin{aligned}
\mathcal{L}_{\mathrm{KD}}(u)
  &= \sum_{x \in \mathcal{U}_t}
     \mathrm{KL}\!\Big(
        p_t^{\mathrm{teach}}(x)\,
        \Big\|\,\sigma\!\big(f(x; u)/T\big)
      \Big) \\
  &\quad + \frac{\mu}{2}\,\|u - w_t\|_2^2 .
\end{aligned}
\end{equation}

    where the first term is the Kullback–Leibler divergence between the teacher’s soft label and the global model’s predicted distribution, and the second term is a small anchor (L2 regularization) that keeps $u$ close to the previous global weights $w_t$. We use a few epochs or gradient steps (e.g. using Adam optimizer) on this objective to obtain updated weights $u_{t+1}$. The anchor term (with coefficient $\mu$) prevents the server update from overshooting or deviating too far due to unreliable logits in any single round. (Notably, this anchored KD step happens entirely on the server; clients do not need to implement or be aware of it.)
    \item \textbf{EMA smoothing and broadcast:} Finally, the server applies an exponential moving average to smooth the global model update. The EMA buffer is updated as $\bar{w}_{t+1} = (1-\beta)u_{t+1} + \beta\bar{w}_t$, where $\beta \in [0,1)$ is the smoothing factor (e.g. $\beta=0.9$). The next round’s model for broadcast is set to $w_{t+1} \leftarrow \bar{w}_{t+1}$. This EMA step acts as a low-pass filter on the sequence of global models, damping the effect of any noisy or biased update. The new global model $w_{t+1}$ is then broadcast to clients at the start of round $t+1$. (If client models are heterogeneous, they may ignore the broadcasted weights and just receive the signal to start the next round using their own last model—our method still works because knowledge is transferred via logits.)
\end{enumerate}

\paragraph{Why EMA is applied after distillation.}
We apply EMA \emph{after} the server distillation step because the KD update is the main source of round-to-round variability (it depends on the sampled clients and their non-IID proxy predictions), and EMA is most effective when it smooths the resulting parameter trajectory. If EMA were applied \emph{before} distillation, it would mainly change the initialization of the KD optimization but would not filter the newly produced update of the current round. Applying EMA \emph{during} distillation would couple temporal smoothing with the inner optimization dynamics, increasing lag and complexity without clear benefit. This “post-update EMA” ordering follows the common practice of maintaining an EMA teacher from the updated student to obtain a more stable target over time \citep{tarvainen2017meanteacher}.

\subsection*{Communication and computation profile.}
\textbf{Uplink:} Each client’s upload per round is only the logits on $Q_k = |\mathcal{U}_{t,k}|$ samples (plus their indices), which is about $2 C Q_k$ bytes (for FP16 probabilities). In our experiments this ranged from 0.09–0.46MB per client per round—orders of magnitude smaller than full model uploads (often tens of MB).\\
\textbf{Downlink:} The server still sends out the full model $w_{t+1}$ each round (which can be partly mitigated by model compression techniques or by sending only model deltas). Note that the downlink size is the same as FedAvg; FedEMA–Distill primarily cuts the uplink cost.\\ \textbf{Client computation:} Clients train locally for $E$ epochs just as in FedAvg. The only extra work is forward-pass inference on $Q_k$ proxy samples to produce logits, which is negligible compared to $E$ epochs on their own data. \\\textbf{Server computation:} The server performs a KD update on at most $|\mathcal{U}_t| = \rho|\mathcal{U}|$ samples each round. In practice this overhead is minor if $|\mathcal{U}|$ is small (e.g. a few thousand points). If needed, the server could also perform KD less frequently (e.g. every few rounds) to reduce computation, while still applying EMA every round.

\subsection{Hyperparameters and practical choices.}
The standard choices work well for FedEMA–Distill, and no extensive hyperparameter tuning was needed.

\subsection{Stability of EMA-guided distillation.}
A key challenge in logits-based FL under strong non-IID data is the \emph{round-to-round variability} of the distilled teacher: in each round only a subset of clients participates, their local models may drift toward different label marginals, and the resulting proxy predictions can be noisy. FedEMA--Distill mitigates this instability through three complementary mechanisms. 
First, the server forms a \emph{consensus teacher} by aggregating client proxy predictions sample-wise, optionally with robust statistics (median/trimmed-mean), which reduces sensitivity to outlier clients and lowers the variance of the teacher signal seen by the server distillation step. Second, after the server KD update, we apply an \emph{exponential moving average} (EMA) on the global weights, which acts as a temporal low-pass filter on the sequence of server updates and dampens oscillations caused by heterogeneous client participation and label skew. Third, a small \emph{anchor term} in the KD objective encourages continuity of the server model across rounds, preventing abrupt jumps when the proxy teacher changes (e.g., due to different sampled clients).
These mechanisms are controlled by a small set of interpretable hyperparameters: the EMA factor \(\beta\) sets the smoothing–lag trade-off; the distillation temperature \(T\)controls the entropy of soft targets (too low yields near one-hot and higher-variance gradients, too high can over-smooth and inject noise); and the anchor weight \(\mu\) prevents overly aggressive server steps while keeping progress. Finally, increasing the proxy shard size \(Q_k\) reduces the variance of aggregated teacher estimates at the cost of slightly higher uplink, which offers a practical stability–communication knob.

\noindent\textbf{Tuning protocol and fairness.}
To ensure a fair comparison, we applied the same conservative tuning budget to FedEMA--Distill and all baselines. Concretely, we performed a small budgeted search on the \emph{training split only} (never using the test set) and selected hyperparameters based on training stability (no divergence) and the average client training objective at a fixed pilot horizon (a short pilot run with the same client sampling and non-IID partition). We then \emph{froze} the selected hyperparameters and reused them unchanged across all datasets and random seeds. This protocol avoids dataset-specific extensive tuning and prevents any test-set leakage. To avoid notation ambiguity, we denote the FedProx proximal coefficient by $\mu_{\text{prox}}$ and the FedEMA--Distill anchor weight by $\mu_{\text{anc}}$ below.

Key settings include:
\begin{itemize}
    \item \textbf{Distillation temperature $T$:} $T=3$–$5$ was effective in our experiments. A higher temperature provides softer knowledge by emphasizing less confident predictions; too low ($T=1$) yields near one-hot targets (losing useful dark knowledge), while too high ($T\gg 1$) over-smooths and adds noise. We use a moderate $T$ ($T= 5$) to balance these factors.
    \item \textbf{EMA factor $\beta$:} $0.9$–$0.95$. A relatively large EMA factor (e.g. 0.9) provides strong smoothing of the global model trajectory without introducing too much lag. We observed that $\beta=0.9$ gives the best stability–accuracy trade-off: $\beta=0$ (no EMA) leads to noisier convergence, while an extreme $\beta=0.99$ can slow down learning significantly.
    \item \textbf{Anchor weight $\mu$:} $10^{-4}$–$10^{-3}$. Including a small anchor in the KD loss (e.g. $\mu = 10^{-4}$) stabilizes training by preventing drastic server-side model jumps. This did not noticeably hurt accuracy, since the anchor is gentle. A too-large anchor (e.g. $\mu = 10^{-2}$) can make progress overly conservative (slower convergence), so we keep $\mu$ in the $10^{-4}$–$10^{-3}$ range.
    \item \textbf{Proxy dataset usage ($\rho$, $Q_k$):} We typically use the full proxy each round ($\rho=1$) in benchmarking experiments. If server load is a concern, one can use a smaller $\rho$ (e.g. 0.5) so that only half the proxy data is distilled each round – the trade-off is slightly more rounds may be needed for convergence. The shard size $Q_k = |\mathcal{U}_{t,k}|$ per client can be adjusted: larger $Q_k$ increases uplink bytes but provides lower-variance ensemble estimates; we found values like $Q_k=100$–$1000$ (depending on $C$) to be reasonable.
    \item \textbf{Initialization and warm-up:} We initialize the EMA buffer as $\bar{w}_0 = w_0$ (so that EMA starts at the initial global model). In early rounds, if the global model is very poor, the client logits might be uninformative. In practice, we found it helpful to do a short “warm-up” phase: e.g. use simple mean aggregation (no robust aggregator) for the first couple of rounds, or even perform a few FedAvg rounds before switching to distillation. This helps the global model and teacher logits reach a reasonable regime, after which FedEMA–Distill can be run as described.
    \item \textbf{Security and privacy:} FedEMA–Distill is compatible with standard FL security measures. Secure aggregation protocols \citep{bonawitz17secureagg,mansouri2023soksecureagg} can be applied to the uploaded logits, so that the server only ever sees aggregated results (e.g. sum of logits) and not any individual client’s logits. This preserves client privacy even if the server is curious. In addition, differential privacy noise can be added to client outputs if desired: techniques from private FL (like DP-SGD or output perturbation \citep{geyer17dpfl,abadi16dp}) can be layered on the logits before aggregation. Exploring the full privacy-utility trade-off is left to future work, but qualitatively, sharing only model outputs (rather than raw data or gradients) already provides a degree of data abstraction that may mitigate some privacy leakage.
\end{itemize}

\noindent\textbf{Method-specific values used in all experiments.}
 Following the above protocol, we use $T=5$, $\beta=0.9$, and $\mu_{\text{anc}}=10^{-4}$ for FedEMA--Distill. For baselines, we use FedAvg with standard settings, FedProx with $\mu_{\text{prox}}=10^{-2}$, FedAvgM with server momentum $m=0.9$, SCAFFOLD with its default control-variate formulation (no additional tuned coefficient beyond the shared optimizer settings), and FedDF with a distillation temperature $T_{\text{DF}}=3$ for server-side distillation on the public proxy dataset.

\section{System Model}

    \begin{figure}
    \centering
    \includegraphics[width=\linewidth]{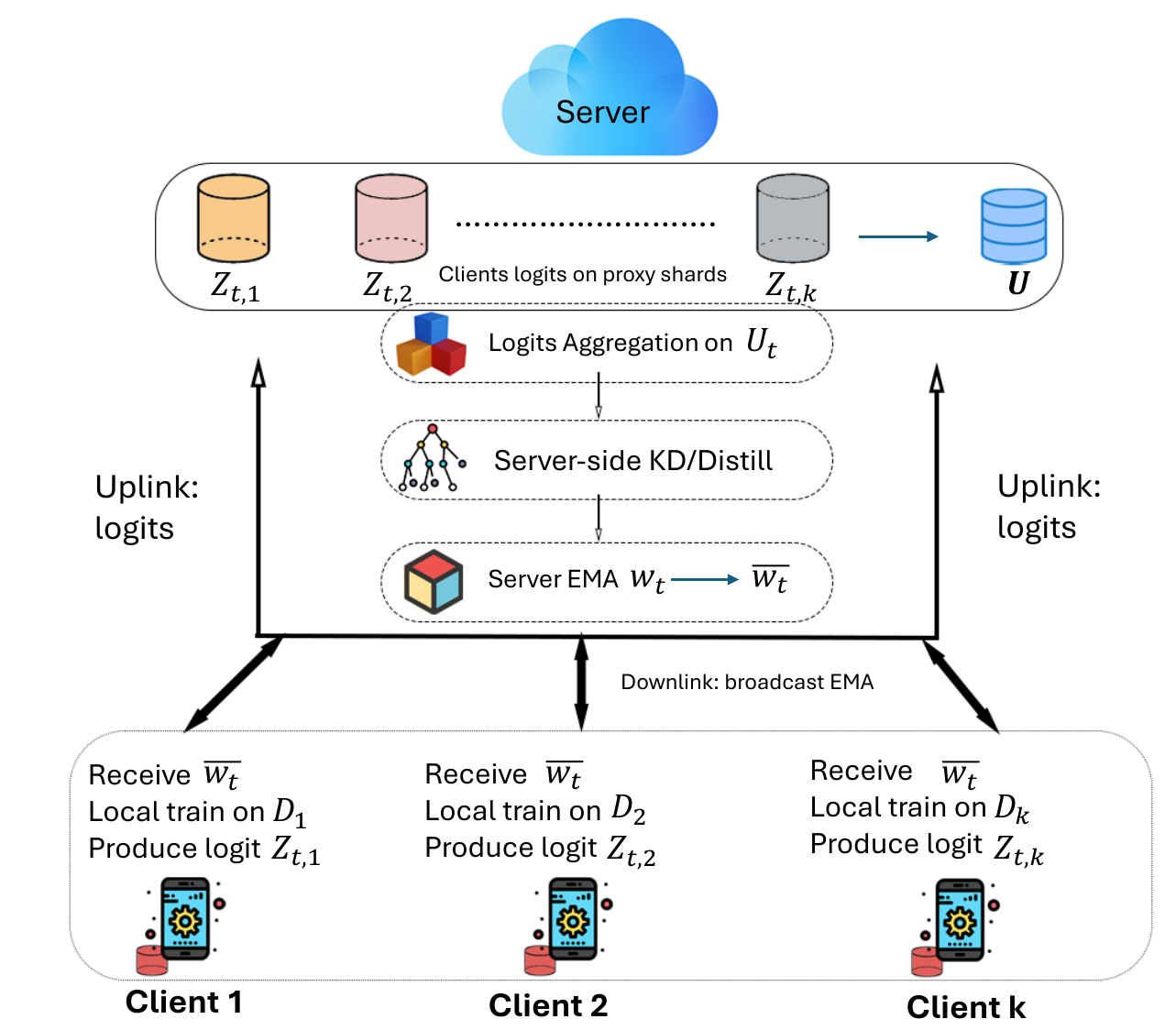}
    \caption{Conceptual overview of FedEMA--Distill.}
    \label{fig:fedema_diagram}
    \end{figure}

We outline entities (server, clients), communication rounds and participation, proxy scheduling, threat model and robustness (Byzantine clients, DP, secure aggregation), and fairness considerations (weighting by $n_k$ if desired). Our protocol is server-driven and requires no client-side code changes beyond computing logits.

    Fig.~\ref{fig:fedema_diagram} provides a round-level diagram of FedEMA--Distill, highlighting logits-only uplink, server-side distillation on the public proxy set, and server EMA smoothing.

\begin{algorithm}
\caption{FedEMA-Distill: server-side round \(t\) with logits-only aggregation and EMA smoothing }
\KwIn{Global weights \(w_t\), EMA buffer \(\bar w_t\); public proxy \(\mathcal{U}\) of size \(U\); participation \(C_{\mathrm{part}}\); local epochs \(E\); temperature \(T\); anchor \(\mu\); EMA factor \(\beta\); coordinate-wise aggregator \(\mathsf{Agg}\) (mean/median/trimmed-mean).}
\KwOut{Updated \((w_{t+1},\,\bar w_{t+1})\).}

Sample client set \(S_t\) with \(|S_t|=\lceil C_{\mathrm{part}}K\rceil\);\quad select \(\mathcal{U}_t\subseteq\mathcal{U}\) and shards \(\{\mathcal{U}_{t,k}\}_{k\in S_t}\).\;
Broadcast \(w_t\) and shard indices to all \(k\in S_t\).\;

\ForEach{client \(k\in S_t\) \textbf{in parallel}}{
  \If{architectures are compatible}{
    \(\theta_{k,t}^{(0)} \leftarrow w_t\)\;
  }
  Train locally for \(E\) epochs on \(D_k\) to obtain \(\theta_{k,t}^{(E)}\).\;
  \ForEach{\(x\in\mathcal{U}_{t,k}\)}{
    \(z_{k,t}(x)\leftarrow f_k(x;\theta_{k,t}^{(E)})\in\mathbb{R}^{C}\);\quad
    \(p_{k,t}(x)\leftarrow \sigma\!\big(z_{k,t}(x)/T\big)\in\Delta^{C-1}\).\;
  }
  Upload FP16 \(\{p_{k,t}(x):x\in\mathcal{U}_{t,k}\}\) (no weights/gradients).\;
}
\ForEach{\(x\in\mathcal{U}_t\)}{ \(\mathcal{K}_t(x)\leftarrow \{k\in S_t:\,x\in\mathcal{U}_{t,k}\}\);
  \(p_t^{\mathrm{teach}}(x)\leftarrow \mathsf{Agg}\big(\{p_{k,t}(x):k\in\mathcal{K}_t(x)\}\big)\)\;
  \textit{(renormalize \(p_t^{\mathrm{teach}}(x)\) to the simplex if median/trimmed-mean is used).}\;
}
\(u_{t+1}\leftarrow w_t\);\quad minimize \(\mathcal{L}_{\mathrm{KD}}(u)\)  on \(\mathcal{U}_t\) to obtain \(u_{t+1}\).\;
\(\bar w_{t+1}\leftarrow (1-\beta)\,u_{t+1}+\beta\,\bar w_t\) \;\(w_{t+1}\leftarrow \bar w_{t+1}\).\;
\Return \((w_{t+1},\,\bar w_{t+1})\).\;
\label{alg:fedema}
\end{algorithm}

\section{Experiments and Results}
We evaluate accuracy, convergence, communication, robustness, calibration, fairness, and ablations.

\subsection{Setup}

We evaluate FedEMA--Distill on four standard federated learning benchmarks, CIFAR-10, CIFAR-100, FEMNIST (via LEAF~\citep{caldas2018leaf}), and AG News~\citep{zhang2015character}, so as to cover both image and text classification tasks with varying numbers of classes and label sparsity. To emulate a challenging yet commonly used non-IID setting, we partition the data across clients using a Dirichlet label distribution with concentration parameter $\alpha = 0.1$, which yields highly skewed client label marginals and is widely adopted as a stress test for FL under strong heterogeneity.

For CIFAR-10/100 we simulate $K = 100$ clients with a per-round participation rate of $C_{\text{part}} = 0.2$, for FEMNIST we use $K = 200$ and $C_{\text{part}} = 0.2$ to reflect its larger natural client population.
For AG News we set $K = 50$ with the same participation rate. As backbone models we adopt ResNet-18 for CIFAR-10/100, a small CNN tailored to character-level inputs for FEMNIST, and a BiGRU with attention mechanism for AG News, which are standard architectures that offer a good trade-off between accuracy and on-device computational cost for their respective modalities. Each selected client performs $E = 5$ local epochs of training. For proxy distillation, the server uses an unlabeled held-out dataset $\mathcal{U}$ with $|\mathcal{U}|$ as reported in Table~\ref{tab:expconfig}, allowing knowledge aggregation without accessing private client data. We compare FedEMA--Distill against representative weight-averaging and distillation baselines: FedAvg~\citep{mcmahan17}, FedProx~\citep{li20fedprox}, SCAFFOLD~\citep{karimireddy20scaffold}, FedAvgM~\citep{hsu19dirichlet}, and FedDF~\citep{lin20feddf},under identical data partitions and a common, budgeted hyperparameter protocol, and report results averaged over three random seeds to reduce variance.

We report all experimental results as \emph{mean $\pm$ standard deviation} over the same three random seeds. When needed, we also compute 95\% confidence intervals (CI) using the Student-$t$ distribution: $\bar{x} \pm t_{0.975,\,S-1}\, (s/\sqrt{S})$ with $S=3$ seeds (so $t_{0.975,2}=4.303$). This provides an uncertainty estimate for both accuracy and efficiency metrics and helps assess the robustness of improvements under non-IID variability.

\paragraph{Train/test split and use of validation data.}
For CIFAR-10/100 and AG News, we use the standard predefined train/test split of each dataset. For FEMNIST, we follow the split provided by the LEAF benchmark. In all cases, the test set is kept strictly held out and is used only for final evaluation and reporting; it is never used for model selection or hyperparameter tuning.

We do not introduce a separate validation set and we do not perform early stopping. Instead, all methods are trained for a fixed number of communication rounds $Q$ (Table~\ref{tab:expconfig}) under the same protocol, and we use fixed hyperparameter choices (no dataset-specific extensive tuning).

From the training split, we additionally sample a disjoint \emph{public proxy} set $\mathcal{U}$ of size $|\mathcal{U}|$ as reported in Table~\ref{tab:expconfig}, which is held out from client training data. The remaining training samples are then partitioned across clients using the Dirichlet label-skew protocol (with $\alpha=0.1$). During knowledge distillation, $\mathcal{U}$ is treated as unlabeled: clients share only soft predictions (logits) on $\mathcal{U}$, and no ground-truth labels from $\mathcal{U}$ are used by FedEMA--Distill.
For fair comparison, for each random seed we fix the sampling of $\mathcal{U}$ and the Dirichlet client partition once, and reuse the exact same split for our method and all baselines.

\begin{table}[t]
\centering
\caption{Key experimental configuration.}
\label{tab:expconfig}
\begin{tabular}{@{}lcccccc@{}}
\toprule
Dataset & $C$ & $K$ & Part. & $E$ & $|\mathcal{U}|$ & $Q$ \\
\midrule
CIFAR-10  & 10  & 100 & 20\% & 5 & 10{,}000 & 500  \\
CIFAR-100 & 100 & 100 & 20\% & 5 & 20{,}000 & 1000 \\
FEMNIST   & 62  & 200 & 20\% & 5 & 10{,}000 & 500  \\
AG News   & 4   & 50  & 20\% & 5 & 5{,}000  & 250  \\
\bottomrule
\end{tabular}
\end{table}

\paragraph{Client-level dataset characteristics}
To better contextualize the reported results under statistical heterogeneity, we summarize the induced client-level data variability for each benchmark.
After removing the public proxy dataset $\mathcal{U}$ (Table~\ref{tab:expconfig}), we construct $K$ clients with a fixed number of local training samples per client (to avoid confounding label skew with client data volume).
Concretely, for each client $k$, we sample class proportions $\boldsymbol{\pi}_k \sim \mathrm{Dir}(\alpha)$ with $\alpha=0.1$ and allocate $n$ samples to client $k$ according to $\boldsymbol{\pi}_k$ (so all clients have the same $n$, while their label composition differs).
In our setup, the resulting samples-per-client are: CIFAR-10 $n=400$, CIFAR-100 $n=300$, FEMNIST $n=3171$, and AG News $n=2300$.

To quantify imbalance and variability, we report in (Table~\ref{tab:clientstats}):
(i) the median number of non-empty classes per client,
(ii) the dominant-class fraction $p_{\max}=\max_c p_{k,c}$ (10th/50th/90th percentiles across clients), and
(iii) the mean normalized label entropy $H_k=-\sum_c p_{k,c}\log p_{k,c}/\log C$ (averaged across clients).
As expected with $\alpha=0.1$, label skew is strongest for low-$C$ tasks (AG News), where many clients concentrate on only a few classes, while high-$C$ tasks exhibit broader but still heterogeneous local label supports.

\begin{table}[t]
\centering
\caption{Client-level data statistics under Dirichlet-$\alpha=0.1$ (after removing $\mathcal{U}$). We report samples/client $n$, median \#non-empty classes per client, dominant-class fraction $p_{\max}$ (P10/P50/P90 across clients), and mean normalized label entropy $H$ (across clients). Statistics are averaged over the three random seeds used in our experiments.}
\label{tab:clientstats}
\small
\begin{tabular}{@{}lcccc@{}}
\toprule
Dataset & $n$ & \#classes/client & $p_{\max}$ & $H$ \\
\midrule
CIFAR-10  & 400  & 5.0  & 0.41/0.67/0.93 & 0.37 \\
CIFAR-100 & 300  & 29.7 & 0.14/0.20/0.31 & 0.59 \\
FEMNIST   & 3171 & 29.0 & 0.17/0.26/0.42 & 0.56 \\
AG News   & 2300 & 2.7  & 0.61/0.90/1.00 & 0.26 \\
\bottomrule
\end{tabular}
\end{table}

\subsection{Main Results: Accuracy and Efficiency}
On CIFAR-10 (Dir-0.1), FedEMA--Distill converges faster (reaches 70\% in $\sim$40 rounds) and to higher final accuracy (80.4\%) than baselines; FedDF (no EMA) is competitive but slightly less stable and lower. Table~\ref{tab:c10} quantifies final accuracy, rounds to 70\%, and per-round uplink.
Across datasets, our method achieves the highest final accuracy (Table~\ref{tab:finalacc}); gains are largest on CIFAR-100. On easier tasks (FEMNIST, AG News), the gain is modest but consistent.

\begin{table}[t]
\centering
\caption{CIFAR-10 (Dir-0.1): final test accuracy, rounds to 70\% and client uplink per round (mean$\pm$std over 3 seeds).}
\label{tab:c10}
\begin{tabular}{@{}lccc@{}}
\toprule
Method & Acc.~(\%)~$\uparrow$ & R@70~$\downarrow$ & Uplink/round \\
\midrule
FedAvg        & $75.2 \pm 0.6$ & $60 \pm 4$ & {3.80}{MB} \\
FedProx       & $77.5 \pm 0.5$ & $55 \pm 3$ & {3.80}{MB} \\
SCAFFOLD      & $77.9 \pm 0.4$ & $51 \pm 3$ & {3.80}{MB} \\
FedAvgM       & $78.3 \pm 0.4$ & $48 \pm 2$ & {3.80}{MB} \\
FedDF         & $79.0 \pm 0.6$ & $42 \pm 2$ & {0.09}{MB} \\
\textbf{FedEMA--Distill} & $\mathbf{80.4 \pm 0.3}$ & $\mathbf{40 \pm 2}$ & \textbf{{0.09}{MB}} \\
\bottomrule
\end{tabular}
\end{table}

\begin{table}[t]
\centering
\caption{Final test accuracy (\%) across datasets (Dir-0.1), reported as mean$\pm$std over 3 seeds.}
\label{tab:finalacc}
\small
\setlength{\tabcolsep}{4pt}
\renewcommand{\arraystretch}{1.05}
\begin{adjustbox}{max width=\columnwidth}
\begin{tabular}{@{}lcccc@{}}
\toprule
Method & CIFAR10 & CIFAR100 & FEMNIST & AG \\
\midrule
FedAvg   & $75.2\!\pm\!0.6$ & $57.0\!\pm\!0.7$ & $85.0\!\pm\!0.2$ & $90.7\!\pm\!0.3$ \\
FedProx  & $77.5\!\pm\!0.5$ & $59.0\!\pm\!0.6$ & $85.5\!\pm\!0.2$ & $90.9\!\pm\!0.3$ \\
SCAFFOLD & $77.9\!\pm\!0.4$ & $60.0\!\pm\!0.5$ & $85.6\!\pm\!0.2$ & $91.0\!\pm\!0.3$ \\
FedAvgM  & $78.3\!\pm\!0.4$ & $60.5\!\pm\!0.5$ & $85.7\!\pm\!0.2$ & $91.1\!\pm\!0.3$ \\
FedDF    & $79.0\!\pm\!0.6$ & $61.5\!\pm\!0.7$ & $85.8\!\pm\!0.3$ & $91.1\!\pm\!0.3$ \\
\textbf{FedEMA-Distill} & $\mathbf{80.4\!\pm\!0.3}$ & $\mathbf{63.0\!\pm\!0.4}$ & $\mathbf{86.3\!\pm\!0.2}$ & $\mathbf{92.0\!\pm\!0.2}$ \\
\bottomrule
\end{tabular}
\end{adjustbox}
\end{table}


\subsection{Communication Efficiency}
We compute total client upload to reach 70\% on CIFAR-10 (Table~\ref{tab:comm70}). Our method needs $\sim${3.6}{MB}/client versus $\sim$ {228}{MB} for FedAvg --- a \emph{63$\times$ reduction}.

\begin{table}[t]
\centering
\caption{CIFAR-10: total client upload to reach 70\% (mean$\pm$std over 3 seeds). Total upload std is derived from rounds std and a constant uplink/round.}
\label{tab:comm70}
\begin{tabular}{@{}lcc@{}}
\toprule
Method & Rounds (mean$\pm$std) & Total upload/client\\
\midrule
FedAvg   & $60 \pm 4$ & ${228.0}\pm{15.2}$\,{MB} \\
FedProx  & $55 \pm 3$ & ${209.0}\pm{11.4}$\,{MB} \\
SCAFFOLD & $51 \pm 3$ & ${193.8}\pm{11.4}$\,{MB} \\
FedAvgM  & $48 \pm 2$ & ${182.4}\pm{7.6}$\,{MB} \\
FedDF    & $42 \pm 2$ & ${3.78}\pm{0.18}$\,{MB} \\
\textbf{FedEMA--Distill} & $\mathbf{40 \pm 2}$ & $\mathbf{{3.60}\pm{0.18}{\,MB}}$ \\
\bottomrule
\end{tabular}
\end{table}

\subsection{Robustness to Byzantine Clients}\label{sec:exp:robust}
\begin{figure}
\centering
\includegraphics[width=\linewidth]{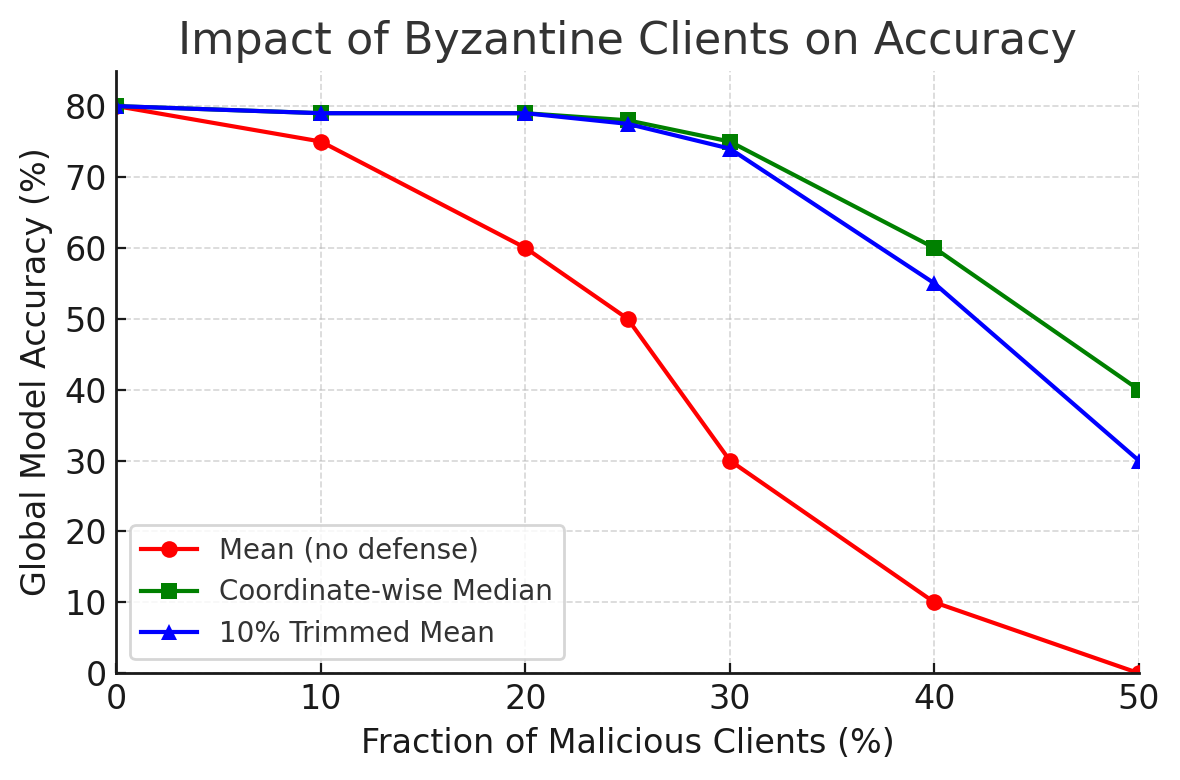}
\caption{Byzantine robustness (CIFAR-10): accuracy vs.\ fraction of malicious clients. Median/trimmed-mean keep performance high up to $\sim$30\% attackers; mean fails early.}
\label{fig:robust}
\end{figure}

We simulate label-flip and random-logit attackers. With naive mean aggregation, performance collapses beyond 10--20\% adversaries; using coordinate-wise median or trimmed mean maintains accuracy up to $\sim$30\% attackers (Figure.~\ref{fig:robust}). Table~\ref{tab:byz25} shows CIFAR-10 final accuracy at 25\% attackers.

\begin{table}[t]
\centering
\caption{CIFAR-10 with 25\% adversaries: final accuracy (\%) by aggregation rule (mean$\pm$std over 3 seeds).}
\label{tab:byz25}
\begin{tabular}{@{}lc@{}}
\toprule
Aggregation & Accuracy (\%) \\
\midrule
Mean (no defense)      & $50.0 \pm 1.5$ \\
Coord.-wise Median     & $78.0 \pm 0.8$ \\
10\% Trimmed Mean      & $77.5 \pm 0.9$ \\
\bottomrule
\end{tabular}
\end{table}

\subsection{Calibration and Fairness}\label{sec:exp:calib-fair}
\begin{figure}
\centering
\includegraphics[width=\linewidth, height=6cm]{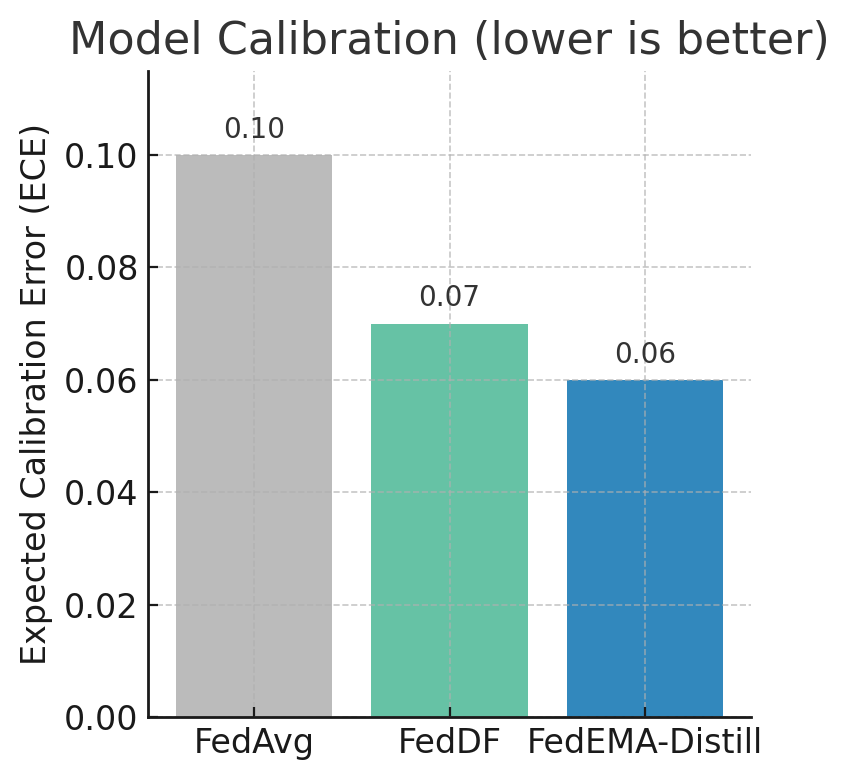}
\caption{Calibration (ECE) on CIFAR-10: lower is better. Distillation and EMA improve calibration.}
\label{fig:ece}
\end{figure}

\textbf{Calibration:} We measure Expected Calibration Error (ECE); FedEMA-Distill attains $\sim$0.06 vs.\ FedAvg’s $\sim$0.10 and FedDF’s $\sim$0.07 (Figure~\ref{fig:ece}).

\begin{figure}
\centering
\includegraphics[width=\linewidth]{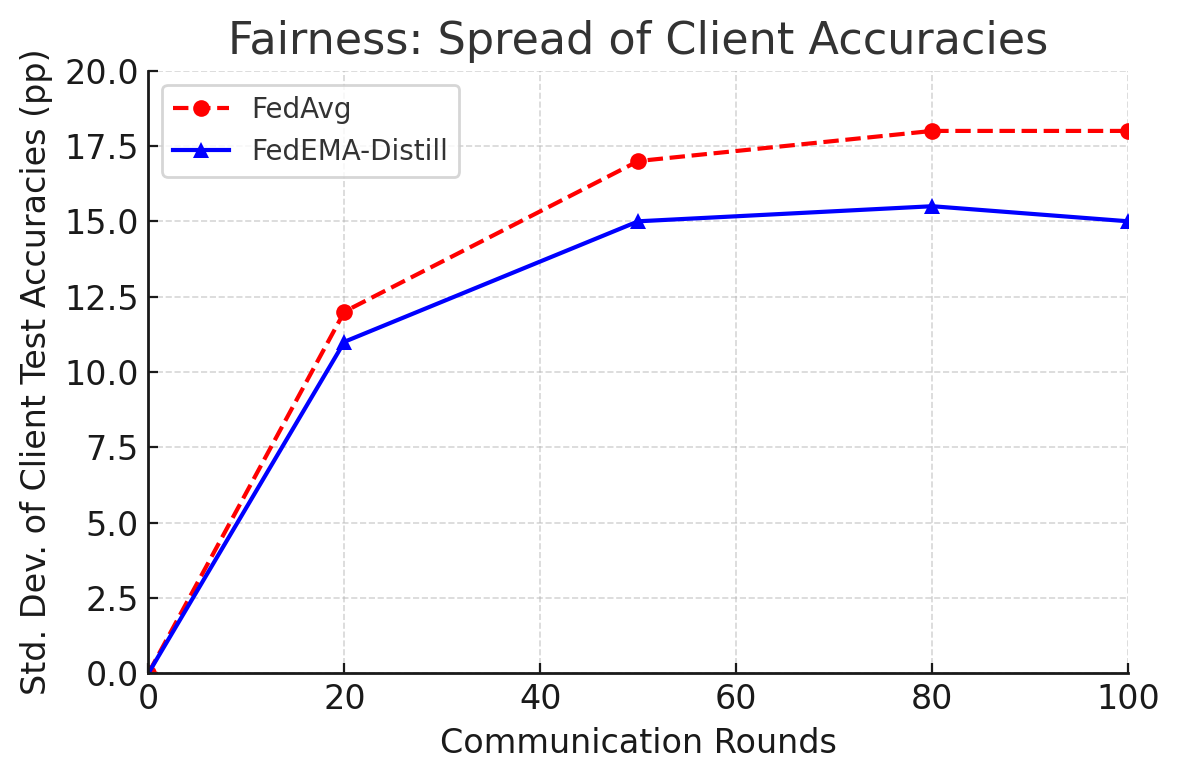}
\caption{Fairness: std.\ dev.\ of per-client accuracies. Lower spread indicates more equitable performance.}
\label{fig:fairness-spread}
\end{figure}

\begin{figure}
\centering
\includegraphics[width=\linewidth]{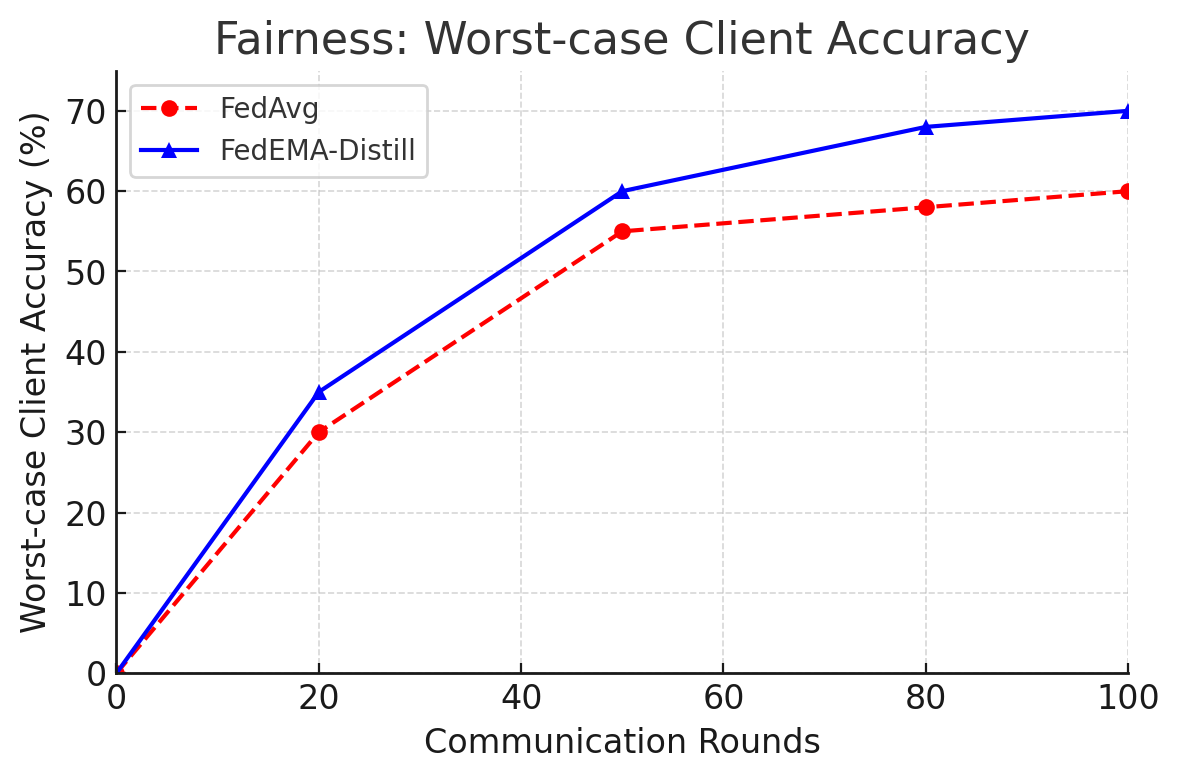}
\caption{Fairness: worst-case client accuracy vs.\ rounds (higher is better).}
\label{fig:fairness-worst}
\end{figure}

\textbf{Fairness:} Our method reduces the std.\ dev.\ of per-client accuracies (e.g., from $\sim$18\,pp to $\sim$15\,pp) and improves worst-case client accuracy by $\sim$10\,pp.
Figure~\ref{fig:fairness-spread} reports the dispersion of client test accuracies (standard deviation across clients) over communication rounds; a lower curve indicates that performance is more evenly distributed instead of being driven by a subset of “easy” clients. We observe that FedEMA--Distill maintains a consistently smaller spread than the baselines, which is consistent with EMA smoothing reducing round-to-round oscillations and limiting client drift under label skew. Complementarily, Figure.~\ref{fig:fairness-worst} tracks the worst-case client accuracy over rounds; the upward shift confirms that the gains are not only average-case, but also improve the tail clients that typically suffer most under strong heterogeneity.

\subsection{Ablations}\label{sec:exp:ablations}
\begin{figure}
\centering
\includegraphics[width=\linewidth]{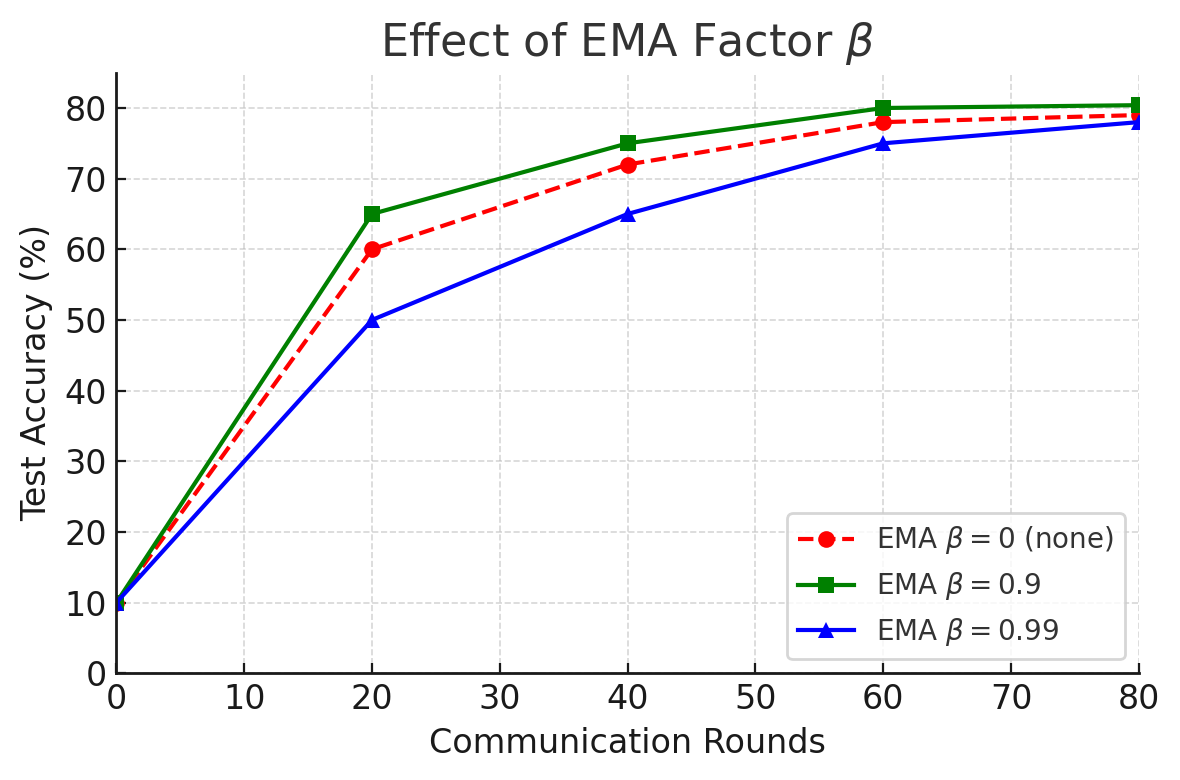}
\caption{Ablation: EMA factor $\beta$.}
\label{fig:beta}
\end{figure}

\textbf{EMA factor $\beta$:} $\beta{=}0$ (no EMA) is noisier and slightly lower final accuracy; $\beta{=}0.9$ best; $\beta{=}0.99$ too sluggish (Figure~\ref{fig:beta}).

\begin{figure}
\centering
\includegraphics[width=\linewidth]{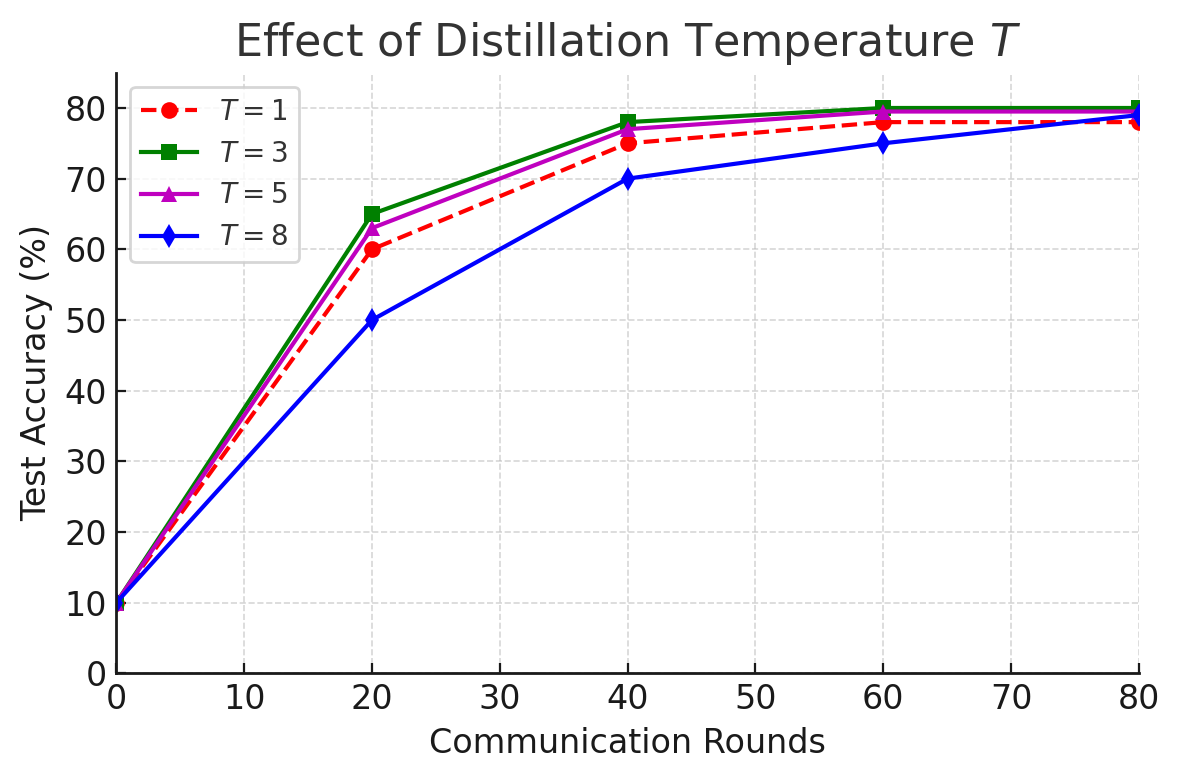}
\caption{Ablation: temperature $T$.}
\label{fig:temp}
\end{figure}

\textbf{Temperature $T$:} $T{=}1$ underperforms; $T{=}3$--$5$ best; $T{=}8$ over-smooths (Figure~\ref{fig:temp}).

 Figure~\ref{fig:beta} visualizes the stability--adaptation trade-off induced by EMA: removing EMA ($\beta=0$) yields a noisier trajectory, while an overly large $\beta$ slows responsiveness to new information, confirming that a moderate EMA factor provides the best balance. Figure~\ref{fig:temp} shows that distillation temperature controls the quality of soft targets: very low $T$ behaves closer to hard labels (less informative and potentially higher-variance gradients), whereas very high $T$ over-smooths the teacher signal; intermediate values ($T=3$--$5$) provide the most stable and accurate convergence in our setting.

\textbf{Proxy usage $\rho,Q$:} halving $Q$ (i.e., $\rho{=}0.5$) preserves final accuracy with $\sim$5 extra rounds.

\textbf{Anchor $\mu$:} small $\mu$ (e.g., $10^{-4}$) stabilizes without slowing; too large slows learning

\noindent\textbf{Discussion (stability interpretation).}
The ablations support the stability role of EMA-guided distillation. Removing EMA (\(\beta=0\)) increases round-to-round fluctuations and leads to noisier convergence, whereas an excessively large\(\beta\) slows adaptation, illustrating the expected smoothing–lag trade-off. Similarly, moderate distillation temperatures(\(T=3\)–\(5\)) provide stable soft targets, while very low \(T\) behaves closer to hard labels and very high \(T\) over-smooths, both of which can degrade the quality and stability of the teacher signal. Together with a small anchor term \(\mu\), these choices explain why coupling EMA with logits-only KD yields a more stable global trajectory under non-IID heterogeneity than distillation-only baselines.

\subsection{Reproducibility Notes}\label{sec:exp:repro}
We fix seeds, average over 3 runs, and implement baselines with matched settings. Hyperparameters for each method are tuned modestly (e.g., FedProx $\mu$, FedAvgM momentum, distillation temperatures).
Hyperparameters follow the original baseline recommendations and the common ranges, we only performed limited sanity-check adjustments on the training split, and the test set was never used for tuning.

\section{Device Energy Analysis}
Using rough energy-per-MB estimates (uplink $\sim$0.25\,J/MB; downlink $\sim$0.15\,J/MB), FedAvg’s $\sim$60 rounds $\times$ {3.8}\,{MB} upload $\approx$ {57}{J}, versus FedEMA-Distill’s $40\times{0.09}{MB}\approx{0.9}$\,J. Downlink also reduces due to fewer rounds. Extra client computation (proxy forwards) is negligible compared to saved communication. Thus, our method is markedly more energy-friendly on devices.

\section{Limitations and Future Work}
While FedEMA--Distill improves stability and communication efficiency through server-side EMA and logits-only distillation, it comes with practical limitations that open clear directions for future work.In real-world federated learning deployments, heterogeneity is the norm rather than the exception: clients typically hold non-IID data due to differences in user behavior, sensors, and local environments, and they may also run different model architectures or capacities because of device constraints. This makes weight-based aggregation less practical when a common architecture cannot be enforced and often amplifies client drift and round-to-round variability under label skew. By exchanging predictions on a shared proxy set instead of full parameters, FedEMA--Distill naturally supports heterogeneous client models and substantially reduces uplink traffic; moreover, the EMA-guided distillation mechanism improves training stability, which can translate into fewer rounds to reach a target accuracy and therefore lower end-to-end communication in bandwidth- and energy-constrained settings.
First, the approach assumes the availability of a \emph{public proxy} dataset $\mathcal{U}$ that is sufficiently representative for the target task. In practice, the effectiveness of logits-based distillation depends on the quality of this proxy: if $\mathcal{U}$ is too small or severely mismatched to the client data distribution, the teacher signal aggregated on $\mathcal{U}$ may become less informative. A natural next step is therefore to reduce reliance on proxy availability by exploring broader choices of proxy sources, including synthetic or generic proxy construction, while preserving the privacy and communication advantages of logits-only exchange.

Second, although server-side distillation is lightweight for the model sizes studied in this paper, scaling the KD step to very large models or extremely large client populations may require additional engineering. In such regimes, the computational cost of repeatedly distilling on $\mathcal{U}$ and aggregating predictions from many clients can become a bottleneck. Practical mitigations include careful proxy subsampling strategies or distributed implementations of the server-side KD pipeline, which we leave for future work.

Third, our communication analysis focuses primarily on uplink cost because it is typically the dominant bottleneck in cross-device FL and is directly affected by switching from weight exchange to logits-only exchange. However, downlink communication can also be significant depending on deployment constraints. Standard downlink compression strategies (e.g., transmitting compact model deltas) are complementary to FedEMA--Distill and could be integrated to further reduce end-to-end communication. Beyond weight-based downlink, an alternative is to broadcast the aggregated teacher logits on the proxy set $\mathcal{U}$, allowing clients to distill locally into their heterogeneous models without downloading a full teacher model each round; exploring this downlink-aware variant and its compute--communication trade-off is an interesting direction for future work.

Fourth, the benefits of aggregation and EMA smoothing can diminish under extremely sparse participation, where very few clients contribute per round and the aggregated proxy teacher becomes highly variable. In such settings, a hybrid strategy that reverts to stronger weight-based aggregation signals when participation is too low may be beneficial. Finally, while robust statistics at the logit level provide protection against simple Byzantine behaviors in our study, more adaptive or sophisticated attacks remain an open challenge. Extending robustness beyond basic defenses (via anomaly detection on uploaded logits) is an important direction for future work.

\section{Conclusion}\label{sec:conclusion}
FedEMA-Distill couples server-side EMA with logits-only distillation, simultaneously improving accuracy/stability under non-IID data and reducing uplink by orders of magnitude. It requires no client changes, supports heterogeneous models, is straightforward to implement, and can be hardened via robust aggregation. Experiments across vision and text tasks validate the approach, with additional benefits for calibration, fairness, and device energy use. We hope this practical design encourages broader deployment of FL in bandwidth- and energy-constrained settings.

\printcredits

\section*{Declaration of competing interest}

The authors declare that they have no known competing financial interests or personal relationships that could have appeared to influence the work reported in this paper.

--

\bibliographystyle{elsarticle-harv}

\bibliography{ref.bib}



\bio{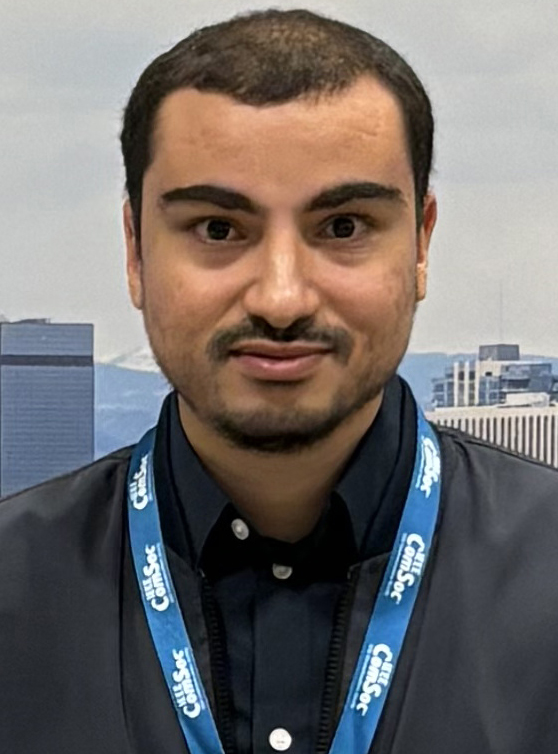}
\textbf{Hamza Reguieg} is a Ph.D. candidate in Engineering Sciences with the Computer Engineering Department, Higher School of Technology (EST), Hassan II University of Casablanca, Morocco, and a member of the Computer Science and Smart Systems (C3S) research unit. He received the Bachelor degree in Mathematics in 2015 and the Master.Sc. degree in Applied Mathematics in 2019 from ENS Casablanca. His research focuses on federated learning and game theory, with an additional interest in internet of things particularly water-resource management. Methodologically, he leverages mathematical modelling and optimization tools, particularly cooperative game theory to design and analyze federated learning algorithms and improve their efficiency and reliability. He has presented and published work at IEEE venues including WINCOM 2023, ICC 2024, and ICC 2025.
\endbio

\bio{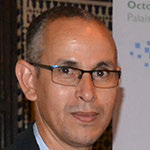}
\textbf{Mohamed El Kamili} received the B.Sc. degree in applied mathematics (statistics) in 1998, the Diplôme d’Études Supérieures Approfondies (D.E.S.A.) degree in numerical analysis, optimization, and operations research in 2000, and the Ph.D. degree in computer science and operations research in 2005, all from Mohammed V University, Rabat, Morocco. He is currently a full-time Professor of applied mathematics with the Computer Engineering Department, Higher School of Technology (EST), Hassan II University of Casablanca, Morocco. He is a permanent member of the research team ``Big Data and Internet of Skills (BDIoS)'' within ``Computer Science and Smart Systems (C3S)''. He has co-authored many journal articles, book chapters, and conference publications. 

His current research interests include networking games; design and evaluation of communication protocols for wireless networks (including wireless MAC protocols); intelligent wireless networks and learning algorithms; cognitive radio and delay-tolerant networks; the IoT, ICN, and D2D communications. From January 2017 to March 2020, he served on the IEEE Morocco Section Committee. He was Local Chair of the 4th edition of WINCOM (Fez, Morocco, October 2016) and Vice-General Chair of the 7th edition of WINCOM (October–November 2019), and has also served as Conference Coordinator. He is a regular reviewer for professional publications and international venues, including \emph{IEEE Network}, \emph{Computer Communications (ComCom)}, IEEE ICC, IEEE GLOBECOM, IEEE IWCMC, IEEE WCNC, and ADHOCNET. 

Prof. El Kamili is the Founder and President of the Moroccan Mobile Computing and Intelligent Embedded-Systems Society (Mobitic; \url{http://www.mobitic.org}) and a co-founder of the International Conference on Wireless Networks and Mobile Communications (WINCOM; \url{http://www.wincom-conf.org}), technically supported by IEEE ComSoc.
\endbio

\bio{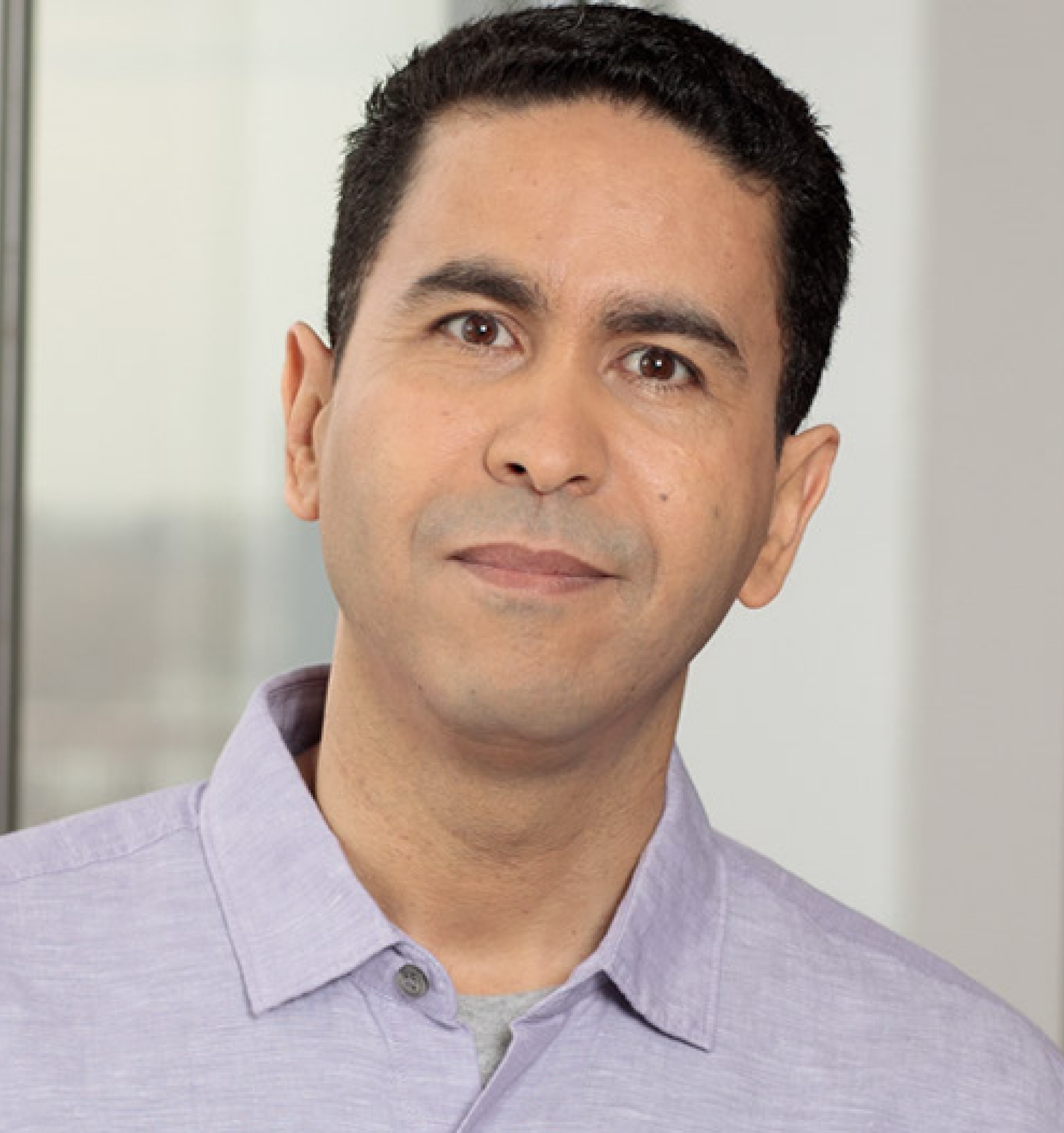} \textbf{Essaid Sabir} (Senior Member, IEEE) received the Ph.D. (Hons.) in Networking and Computer Engineering from Avignon University, France, in 2010. From 2009 to 2024, he held academic positions at Avignon University, Hassan II University of Casablanca, and Université du Québec à Montréal. Since 2024, he has been a full Professor at TÉLUQ, Université du Québec. His research focuses on 5G/6G, IoT, ubiquitous networking, AI/ML, and game theory. He has led several national and international projects and received multiple awards. He actively contributes to the research community as the founder of the UNet conference, a co-founder of WINCOM, an area/associate/guest editor for many top-tier journals, and an organizer of major venues.
\endbio

\end{document}